\documentclass{article}

\usepackage{arxiv}

\usepackage[utf8]{inputenc} 
\usepackage[T1]{fontenc}    
\usepackage{hyperref}       
\usepackage{url}            
\usepackage{booktabs}       
\usepackage{amsfonts}       
\usepackage{nicefrac}       
\usepackage{microtype}      
\usepackage{lipsum}		
\usepackage{graphicx}
\usepackage{natbib}
\usepackage{doi}
\usepackage{amssymb}
\usepackage{amsmath} 
\usepackage{algorithm}
\usepackage{algpseudocode}
\usepackage{booktabs}

\title{CIVIC: End-to-End Sequence Compactness for Efficient Vision-Language Models}

\author{ \href{https://orcid.org/0009-0000-8734-5498}{\includegraphics[scale=0.06]{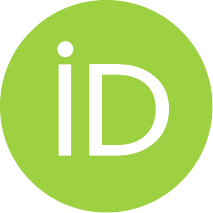}\hspace{1mm}Fengze~Yang} \\
	Department of Civil \& Environmental Engineering\\
	University of Utah\\
	201 Presidents' Cir, \\
    Salt Lake City, UT 84112, USA \\
	\texttt{fred.yang@utah.edu} \\
    \And
	\href{https://orcid.org/0009-0006-7737-1178}{\includegraphics[scale=0.06]{orcid.pdf}\hspace{1mm}Xuewen~Luo} \\
	Department of Civil \& Environmental Engineering\\
	University of Utah\\
	201 Presidents' Cir, \\
    Salt Lake City, UT 84112, USA \\
	\texttt{xuewen.luo@utah.edu} \\
    \And
    \href{https://orcid.org/0009-0003-0211-9953}{\includegraphics[scale=0.06]{orcid.pdf}\hspace{1mm}Bo~Yu} \\
	Department of Civil \& Environmental Engineering\\
	University of Utah\\
	201 Presidents' Cir, \\
    Salt Lake City, UT 84112, USA \\
	\texttt{bo.yu@utah.edu} \\
    \And
	\href{https://orcid.org/0000-0002-5162-891X}{\includegraphics[scale=0.06]{orcid.pdf}\hspace{1mm}Xiaoyue~Cathy~Liu} \\
	Department of Civil \& Environmental Engineering\\
	University of Utah\\
	201 Presidents' Cir, \\
    Salt Lake City, UT 84112, USA \\
	\texttt{cathy.liu@utah.edu} \\
    \And
	\href{https://orcid.org/0000-0002-5447-4768}{\includegraphics[scale=0.06]{orcid.pdf}\hspace{1mm}Chenxi~Liu}\thanks{Corresponding author.} \\
	Department of Civil \& Environmental Engineering\\
	University of Utah\\
	201 Presidents' Cir, \\
    Salt Lake City, UT 84112, USA \\
	\texttt{chenxi.liu@utah.edu} \\
}

\renewcommand{\shorttitle}{\textit{arXiv} Template}

\hypersetup{
pdftitle={A template for the arxiv style},
pdfsubject={q-bio.NC, q-bio.QM},
pdfauthor={David S.~Hippocampus, Elias D.~Striatum},
pdfkeywords={First keyword, Second keyword, More},
}

\begin{document}
\maketitle

\begin{abstract}
	Vision-Language Models (VLMs) face severe memory and latency bottlenecks due to high-resolution visual tokens. While current token reduction methods theoretically save FLOPs, post-hoc pruning introduces structural overhead, failing to yield proportional wall-clock acceleration. However, enforcing a contiguous compact pathway risks geometric disorientation and loss of fine-grained localization. To overcome these barriers, this paper introduces CIVIC, a path-consistent compact visual inference framework. By maintaining compact sequence representations seamlessly across the vision encoder, projection layer, LLM prefill, and KV-cache, CIVIC avoids non-contiguous memory access and localized unmerging overheads. Evaluated on the Qwen3-VL architecture, CIVIC successfully translates sequence reductions into genuine physical hardware efficiency, shrinking KV-cache memory to approximately one-third of the baseline and reducing end-to-end inference latency. Enabled by text-aligned KL distillation and an adaptive spatial retention floor, CIVIC achieves these efficiency milestones without degrading accuracy across rigorous multimodal reasoning and visual grounding benchmarks.
\end{abstract}


\section{Introduction}

Vision-language models (VLMs) have become an important paradigm for multimodal reasoning by integrating visual perception and language understanding within a unified framework for multimodal reasoning. Applied to video understanding, robotics, and autonomous driving, modern VLMs process observations into semantically grounded outputs \citep{khaki2025sparsevila, yu2026visiontrim}. However, relying on high-resolution images or long videos generates massive visual token sets and extended contexts \citep{yang2025visionzip, hu2025lightvlm}. Because transformer attention scales with sequence length, dense representations drastically increase computation, KV cache footprints, memory use, and latency \citep{tu2024vlcache, zeng2026hybridkv}. This severely bottlenecks resource-constrained, low-latency deployments, making efficient VLM inference a critical research challenge for practical distribution.

Recent studies have shown that substantial visual redundancy exists in VLM pipelines, motivating extensive research on token sparsification, token merging, and adaptive visual compression. Existing methods reduce redundant computation by pruning low-importance tokens, dynamically sparsifying visual representations, or merging semantically similar patches during inference \citep{rao2021dynamicvit, bolya2022tome, chen2023diffrate, wang2025dymu, yang2025visionzip, khaki2025sparsevila, yang2025vflowopt, yu2026visiontrim, kim2025training}. Another line of research focuses on multimodal decoding efficiency through KV-cache compression, compact-prefill strategies, and modality-aware cache management to reduce long-context inference costs \citep{tu2024vlcache, zeng2026hybridkv, li2026packcache, huang2025aircache, hu2025lightvlm}. Meanwhile, compact visual projectors, latent token compression, and efficient multimodal backbones have further encouraged the use of compressed multimodal representations for scalable inference \citep{lee2024learning, li2025tokenpacker, lu2025internvlx, zhang2025llavamini, wang2025leomini, wang2025flash}. These studies collectively indicate a growing trend toward compact multimodal representations and latent-space reasoning for efficient VLM inference.

\begin{figure}[h]
  \centering
  \includegraphics[width=0.75\columnwidth]{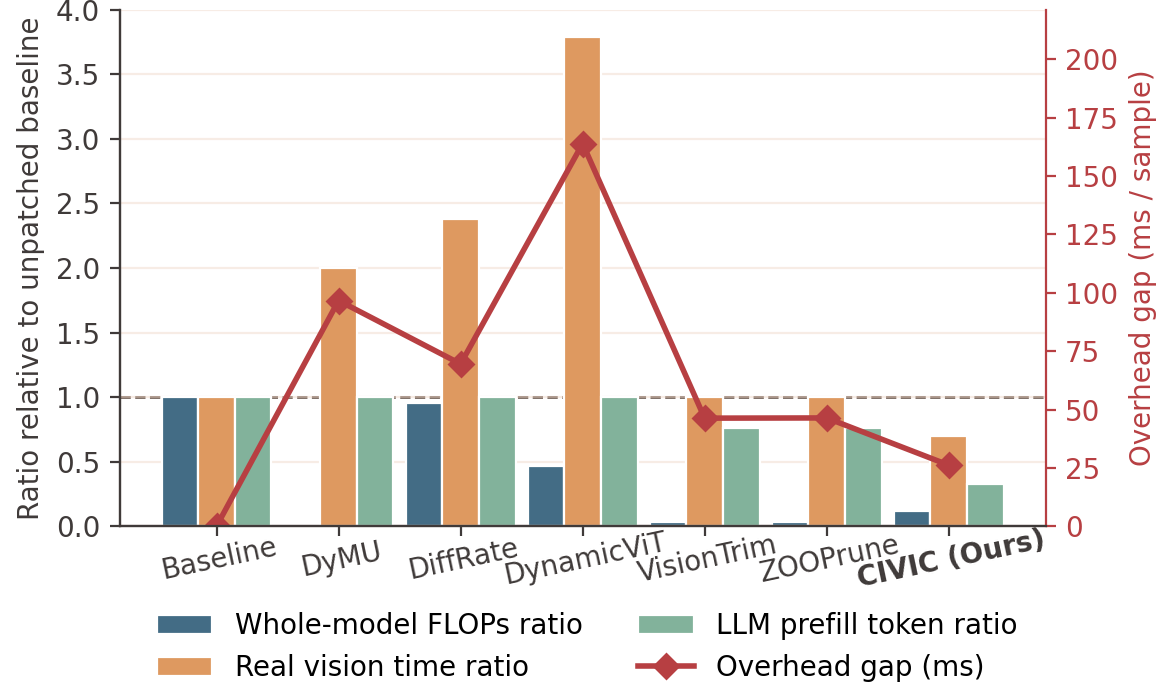}
  \caption{Comparison between theoretical compression and practical inference behavior across recent efficient VLM methods on Qwen3-VL 2B model.}
  \label{fig:gap_analysis_summary}
\end{figure}

Despite recent progress, reducing estimated visual FLOPs does not always translate into proportional inference acceleration in modern VLMs. As illustrated in Fig.~\ref{fig:gap_analysis_summary}, recent methods effectively reduce visual token counts and theoretical computation, demonstrating the importance of compact multimodal representations. However, practical deployment can still be affected by additional runtime operations, compatibility restoration, and different VLM structure. In many cases, compact representations primarily function as intermediate compression states, while downstream inference continues to rely on partially dense multimodal interfaces. These challenges become more evident when the compression methods are applied on VLMs that were not designed for them. These suggesting that efficient VLM inference depends on a stable and general compact processing pathways throughout multimodal inference.

To address these challenges, this paper proposes \textbf{C}ompact \textbf{I}nference for \textbf{V}ision-Language \textbf{I}ntegrated \textbf{C}ompression (CIVIC). However, enforcing a contiguous compact pathway presents inherent challenges: disrupting 2D grids causes geometric disorientation, over-compression destroys fine-grained localization, and altering tensor dimensions breaks traditional distillation. To overcomes these barriers, CIVIC enforces a path-consistent design where compact latents serve as the primary inference pathway. Specifically, dense patches are converted into consecutive tokens via learned anchor-based aggregation to maintain spatial mapping. The sequence is then processed using KV-compressed attention, regulated by an adaptive spatial retention floor to preserve localizing details. Finally, a text-aligned KL distillation scheme bypasses structural training boundaries, allowing compact embeddings to directly replace dense placeholder spans in the LLM prefill. This end-to-end compactness maximizes physical hardware efficiency without the overhead of repeated dense restorations or dynamic compatibility reconstructions.

Extensive experiments on Qwen3-VL show that CIVIC maintains compact representations throughout multimodal inference, enabling a more stable and implementation-friendly compact processing pathway for efficient VLM deployment. The main contributions of this paper are summarized as follows:
\begin{itemize}

    \item This paper identifies the compression-realization gap in VLM inference: theoretical FLOP reductions fail to yield physical acceleration unless sequence compactness is preserved across the vision encoder, projector, and LLM prefill.

    \item This paper proposes CIVIC, a path-consistent framework that resolves geometric and training bottlenecks by utilizing learned anchor-based aggregation, an adaptive spatial retention floor, and text-aligned KL distillation to maintain end-to-end compact latent representations.

    \item This paper evaluates CIVIC on Qwen3-VL and shows that stable compact processing improves practical deployment-oriented efficiency—drastically reducing latency and KV-cache footprints—while preserving fine-grained multimodal accuracy and eliminating reliance on runtime routing or dense restoration.

\end{itemize}

\section{Related Work}

\subsection{Visual Token Compression.}
A broad class of efficient VLM methods reduces the visual sequence before or during multimodal inference. Earlier vision-transformer studies explored adaptive token dropping, token merging, and learnable compression schedules to reduce redundant computation in dense visual encoders \citep{rao2021dynamicvit, bolya2022tome, chen2023diffrate, lee2024learning}. Recent multimodal methods adapt these ideas to VLM pipelines by selecting visually dominant tokens, preserving text-relevant regions, compressing visual information flow, or estimating token sensitivity during inference \citep{wang2025dymu, yang2025visionzip, khaki2025sparsevila, yang2025vflowopt, yu2026visiontrim, kim2025training}. These methods show that visual redundancy is substantial and that compact token sets can preserve much of the original multimodal capability. However, many approaches rely on runtime scoring, token routing, gather/scatter operations, or restoration mechanisms, which can limit practical acceleration in complex VLM systems.

\subsection{Compact Multimodal Representations.}
Another direction redesigns the interface between visual encoders and language models. TokenPacker reduces visual token redundancy through an efficient projector \citep{li2025tokenpacker}. InternVL-X incorporates visual token compression into the InternVL family through projector-level, layer-wise, and resolution-aware compression \citep{lu2025internvlx}. LLaVA-Mini explores extreme compact-prefill by using one vision token after modality pre-fusion \citep{zhang2025llavamini}. LEO-MINI combines conditional token reduction with multimodal experts \citep{wang2025leomini}, while FLASH studies latent-aware multimodal decoding \citep{wang2025flash}. These studies indicate a broader trend toward compact multimodal representations rather than treating efficiency only as post-hoc pruning.

\subsection{KV-Cache and Prefill Optimization.}
Long multimodal contexts also increase the memory and computation required by language-model prefill and decoding. VL-Cache \cite{tu2024vlcache}, AirCache \cite{huang2025aircache}, HybridKV \cite{zeng2026hybridkv}, and PackCache \cite{li2026packcache} compress multimodal cache states to reduce memory and decoding overhead. LightVLM combines visual token merging with KV-cache compression, showing that visual compression and cache management can be jointly considered \citep{hu2025lightvlm}. These studies show that efficient VLM deployment requires reducing not only visual encoder computation but also the sequence and cache burden seen by the LLM.

CIVIC advances compact multimodal inference by enforcing persistent compact processing across VLM pipelines. Rather than treating compact tokens as intermediate states, CIVIC compresses dense patches before the vision transformer, processes them using KV-compressed attention, and inserts the resulting embeddings directly into the language model prefill. This end-to-end design eliminates runtime routing and dense-restoration overhead while preserving compatibility with existing VLM generation interfaces.

\section{Methodology}
\label{sec:methodology}

\subsection{Problem Formulation}
\label{subsec:problem_formulation}

Let $\mathcal{F}$ denote a visual input, $x=\{x_i\}_{i=1}^{L}$ denote the textual prompt, and $\hat{y}$ denote the generated output. Let $T_e$ and $D_v$ denote the dense visual token length and visual hidden dimension inside the visual encoder, and let $T_p$ and $D_l$ denote the number of visual tokens inserted into the language-model prefill and the language hidden dimension. A dense VLM inference pipeline is defined in Eq.~\ref{eq:dense_vlm_pipeline}, where $E_v$ is the visual encoder, $\Pi$ is the multimodal projector, $\mathrm{Merge}(\cdot)$ inserts projected visual tokens into the language-model context, and $\mathrm{LM}_{\theta}$ is the autoregressive language model.
\begin{align}
V &= \Pi(E_v(\mathcal{F})),
\qquad
V\in\mathbb{R}^{T_p\times D_l},
\nonumber\\
\tilde{x} &= \mathrm{Merge}(x,V),
\qquad
|\tilde{x}|=L+T_p,
\nonumber\\
\hat{y} &= \mathrm{LM}_{\theta}(\tilde{x}).
\label{eq:dense_vlm_pipeline}
\end{align}

Let $N_v$ and $N_l$ denote the numbers of visual and language layers. The dense inference cost $\mathcal{C}_{\mathrm{dense}}$ is approximated in Eq.~\ref{eq:dense_cost}, where $\alpha$, $\beta$, and $\gamma$ are implementation-dependent constants, and $\mathcal{C}_{\mathrm{dec}}$ denotes autoregressive decoding cost.
\begin{align}
\mathcal{C}_{\mathrm{dense}}
&=
\underbrace{\alpha N_v T_e^2D_v}_{\text{visual self-attention}}
+\;
\underbrace{\beta N_l(L+T_p)^2D_l}_{\text{LLM prefill attention}}
\nonumber\\
&\quad+\;
\underbrace{\gamma N_l(L+T_p)D_l}_{\text{KV-cache storage}}
\;+\;\mathcal{C}_{\mathrm{dec}} .
\label{eq:dense_cost}
\end{align}

Let $\mathcal{R}$ denote a post-hoc runtime compression procedure applied after dense visual-token construction, and let $\mathcal{O}_{\mathcal{R}}$ denote its induced runtime operation set, including token scoring, selection, gather/scatter, merge/unmerge, and dense-interface restoration. The routing overhead is defined in Eq.~\ref{eq:route_cost}, where $c(o)$ denotes the runtime cost of operation $o$.
\begin{equation}
\mathcal{C}_{\mathrm{route}}(\mathcal{R})
=
\sum_{o\in\mathcal{O}_{\mathcal{R}}} c(o).
\label{eq:route_cost}
\end{equation}

The practical deployment cost for a post-hoc compression method is represented in Eq.~\ref{eq:posthoc_compression}.
\begin{align}
\Omega_{\mathrm{post}}
&=
\mathcal{C}_{\mathrm{dense}}
+
\mathcal{C}_{\mathrm{route}}(\mathcal{R}) .
\label{eq:posthoc_compression}
\end{align}

In contrast, path-consistent compact visual inference is defined in Eq.~\ref{eq:native_compact_def}. Let $X_0\in\mathbb{R}^{T_e\times D_v}$ denote dense patch embeddings before visual encoding, $\mathcal{A}$ denote the compact aggregation function, $E_v^c$ denote the compact visual encoder, and $\Pi_c$ denote the compact multimodal projector. The compact visual length inserted into the language-model prefill is $M_p$, where $M_p<T_p$.
\begin{align}
V_c &= \Pi_c(E_v^{c}(\mathcal{A}(X_0))),
\;
V_c\in\mathbb{R}^{M_p\times D_l},
\nonumber\\
\tilde{x}_c &= \mathrm{Merge}(x,V_c),
\;
|\tilde{x}_c|=L+M_p .
\label{eq:native_compact_def}
\end{align}

\subsection{Overview of CIVIC}
\label{subsec:civic_overview}

\begin{figure*}[h]
    \centering
    \includegraphics[width=1\linewidth]{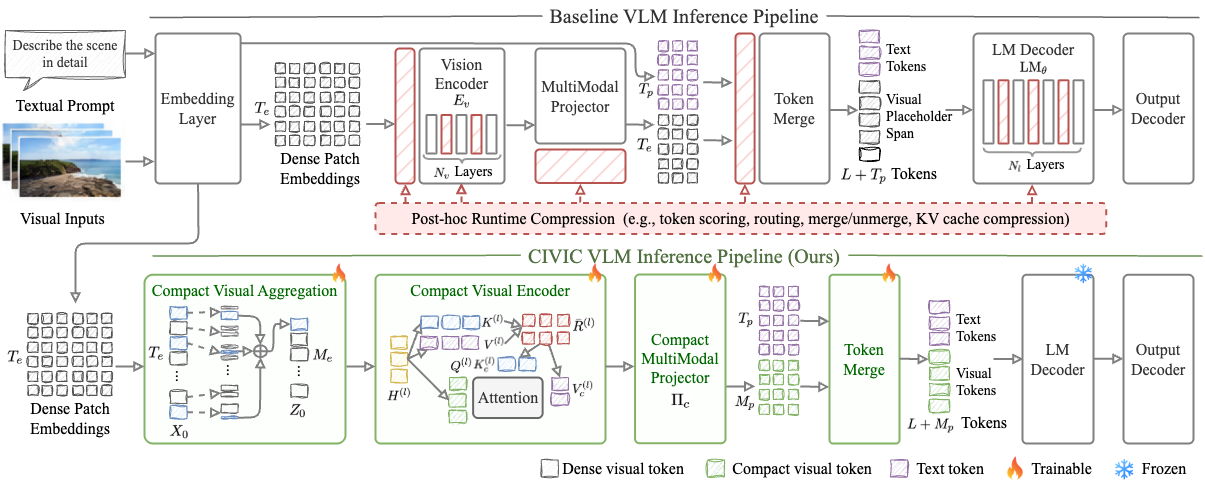}
    \caption{Overview of the proposed CIVIC pipeline. Existing methods typically apply post-hoc runtime compression after dense visual representations have already been constructed, requiring additional operations across the multimodal pipeline. In contrast, CIVIC performs compact visual aggregation before visual encoding and preserves compact latent representations throughout following inference.}
    \label{fig:method}
\end{figure*}

As shown in Fig.~\ref{fig:method}, CIVIC defines a compact inference function $G_{\theta,\phi}^{c}$ by replacing the dense visual pathway in Eq.~\ref{eq:dense_vlm_pipeline} with the native compact pathway in Eq.~\ref{eq:native_compact_def}. Here, $\theta$ denotes frozen VLM parameters and $\phi$ denotes CIVIC-specific parameters. The optimization objective is given in Eq.~\ref{eq:civic_obj}, where $\mathcal{L}$ denotes the distillation loss and $\Omega(\cdot)$ denotes deployment cost.
\begin{align}
\phi^{*}
&=
\arg\min_{\phi}\;
\mathbb{E}_{(\mathcal{F},x)}
\left[
\mathcal{L}
\left(
G_{\theta}(\mathcal{F},x),
G_{\theta,\phi}^{c}(\mathcal{F},x)
\right)
\right],
\nonumber\\
&\text{s.t.}\qquad
\Omega(G_{\theta,\phi}^{c})\leq \Omega_{\max}.
\label{eq:civic_obj}
\end{align}

CIVIC reduces the encoder-side visual length from $T_e$ to $M_e$ and the prefill-side visual length from $T_p$ to $M_p$. Let $S$ denote the number of key/value anchors used by compact visual attention. The compact model cost is approximated in Eq.~\ref{eq:civic_model_cost}.
\begin{align}
\mathcal{C}_{\mathrm{model}}^{c}
&=
\underbrace{\alpha N_v M_eSD_v}_{\text{compact visual attention}}
+
\underbrace{\beta N_l(L+M_p)^2D_l}_{\text{compact LLM prefill}}
\nonumber\\
&\quad+
\underbrace{\gamma N_l(L+M_p)D_l}_{\text{compact KV-cache storage}}
+\;
\mathcal{C}_{\mathrm{dec}} .
\label{eq:civic_model_cost}
\end{align}

Compared with Eq.~\ref{eq:dense_cost}, Eq.~\ref{eq:civic_model_cost} changes the visual attention term from $T_e^2$ to $M_eS$ and changes the language-model prefill length from $L+T_p$ to $L+M_p$. Compared with Eq.~\ref{eq:posthoc_compression}, CIVIC avoids repeated oprations in $\mathcal{O}_{\mathcal{R}}$ by with path-consistent compact visual inference.

\subsection{Compact Visual Aggregation}
\label{subsec:anchor_aggregation}

Let $X_0\in\mathbb{R}^{T_e\times D_v}$ be dense patch embeddings before visual encoding, and let $A\in\mathbb{R}^{M_e\times D_v}$ be learnable compact anchors. CIVIC defines the aggregation function $\mathcal{A}$ in Eq.~\ref{eq:anchor_aggregation}, where $\mathrm{LN}(\cdot)$ is layer normalization, $\mathrm{Norm}(\cdot)$ is $\ell_2$ normalization, $\tau$ is the aggregation temperature, and the softmax is applied over the dense-token dimension.
\begin{align}
K_0 &= \mathrm{Norm}(\mathrm{LN}(X_0)),
\;\;
Q_A = \mathrm{Norm}(\mathrm{LN}(A)),
\nonumber\\
W_A &= \mathrm{softmax}(Q_AK_0^{\top}/\tau),
\nonumber\\
Z_0 &= \mathcal{A}(X_0)=W_AX_0,
\;\;
Z_0\in\mathbb{R}^{M_e\times D_v}.
\label{eq:anchor_aggregation}
\end{align}

\subsection{Compact Visual Encoding}
\label{subsec:kv_compressed_attention}

The compact visual encoder $E_v^c$ processes $Z_0$ instead of the dense sequence $X_0$. Let $H^{(l)}\in\mathbb{R}^{M_e\times D_v}$ be compact hidden states at visual layer $l$. CIVIC uses the original query projection while compressing keys and values into $S$ memory anchors. The layer computation is defined in Eq.~\ref{eq:compact_visual_layer}, where $W_Q^{(l)}$, $W_K^{(l)}$, $W_V^{(l)}$ are projection matrices, $R^{(l)}\in\mathbb{R}^{M_e\times S}$ is the KV assignment matrix, $\bar{R}^{(l)}$ is column-normalized, and $d$ is the per-head dimension.
\begin{align}
Q^{(l)} &= H^{(l)}W_Q^{(l)},
\;
K^{(l)} = H^{(l)}W_K^{(l)},
\;
\nonumber\\
V^{(l)} &= H^{(l)}W_V^{(l)},
\nonumber\\
K_c^{(l)} &= {\bar{R}^{(l)\top}}K^{(l)},
\;
V_c^{(l)} = {\bar{R}^{(l)\top}}V^{(l)},
\nonumber\\
\mathrm{Attn}_{c}^{(l)}
&=
\mathrm{softmax}
\left(
\frac{Q^{(l)}{K_c^{(l)\top}}}{\sqrt{d}}
\right)
V_c^{(l)} .
\label{eq:compact_visual_layer}
\end{align}

Eq.~\ref{eq:compact_visual_layer} changes the attention score size from $M_e\times M_e$ to $M_e\times S$. Relative to dense visual attention over $T_e$ tokens, the per-layer attention interaction ratio is approximated in Eq.~\ref{eq:attention_ratio}.
\begin{equation}
\frac{\mathcal{C}_{\mathrm{attn}}^{c}}
{\mathcal{C}_{\mathrm{attn}}^{d}}
\approx
\frac{M_eS}{T_e^2}.
\label{eq:attention_ratio}
\end{equation}

The compact encoder output is denoted in Eq.~\ref{eq:compact_encoder_output}.
\begin{equation}
H_c = E_v^c(Z_0,P_c),
\qquad
H_c\in\mathbb{R}^{M_e\times D_v}.
\label{eq:compact_encoder_output}
\end{equation}

\subsection{Compact Multimodal Prefill}
\label{subsec:compact_prefill}

The compact multimodal projector $\Pi_c$ maps $H_c$ into the language hidden space. If a VLM contains a visual merger before projection, it is applied to the compact sequence rather than to the dense sequence. The projected compact visual embeddings are defined in Eq.~\ref{eq:compact_visual_projection}.
\begin{equation}
V_c=\Pi_c(H_c),
\;
V_c\in\mathbb{R}^{M_p\times D_l},
\;
M_p<T_p.
\label{eq:compact_visual_projection}
\end{equation}

Let $E_x\in\mathbb{R}^{L\times D_l}$ be text embeddings. CIVIC replaces dense visual placeholder spans with $V_c$ and constructs the compact multimodal prefill in Eq.~\ref{eq:compact_prefill_sequence}.
\begin{equation}
\tilde{x}_{c}
=
\mathrm{Merge}(E_x,V_c),
\;
|\tilde{x}_{c}|=L+M_p .
\label{eq:compact_prefill_sequence}
\end{equation}

The resulting prefill and KV-cache ratios relative to dense inference are given in Eq.~\ref{eq:prefill_ratio}.
\begin{equation}
\frac{\mathcal{C}_{\mathrm{prefill}}^{c}}
{\mathcal{C}_{\mathrm{prefill}}^{d}}
\approx
\frac{(L+M_p)^2}{(L+T_p)^2},
\;
\frac{\mathcal{C}_{\mathrm{cache}}^{c}}
{\mathcal{C}_{\mathrm{cache}}^{d}}
\approx
\frac{L+M_p}{L+T_p}.
\label{eq:prefill_ratio}
\end{equation}

Text positions are preserved. Compact visual positions are inherited from the dense visual span. Intermediate visual features required by the model interface are produced from the compact stream and indexed by compact visual segments.

\subsection{Model Training}
\label{subsec:distillation}

Distillation is used for parameter training because CIVIC changes the visual-token interface while preserving the pretrained language model. Let $G_{\theta}$ be the dense teacher and $G_{\theta,\phi}^{c}$ be the compact student. The teacher and student logits are denoted as $\ell_t$ and $\ell_s$. Since dense and compact visual placeholder positions do not align one-to-one, distillation is computed over aligned nonvisual text positions. Let $\mathcal{I}_t$ and $\mathcal{I}_s$ denote teacher and student text-position sets, and let $\pi:\mathcal{I}_t\rightarrow\mathcal{I}_s$ map teacher positions to corresponding student positions.

The temperature-scaled distributions and text-aligned KL loss are defined in Eq.~\ref{eq:distillation_loss}, where $T_{\mathrm{KL}}$ is the distillation temperature and $\lambda$ controls the loss weight.
\begin{align}
p_t^{i}
&=
\mathrm{softmax}
\left(
\ell_t^i/T_{\mathrm{KL}}
\right),
\;
i\in\mathcal{I}_t,
\nonumber\\
p_s^{\pi(i)}
&=
\mathrm{softmax}
\left(
\ell_s^{\pi(i)}/T_{\mathrm{KL}}
\right),
\;
\pi(i)\in\mathcal{I}_s,
\nonumber\\
\mathcal{L}_{\mathrm{KL}}
&=
\frac{\lambda T_{\mathrm{KL}}^2}{|\mathcal{I}_t|}
\sum_{i\in\mathcal{I}_t}
\mathrm{KL}
\left(
p_t^i
\;\|\;
p_s^{\pi(i)}
\right),
\nonumber\\
\phi^{*}
&=
\arg\min_{\phi}
\mathcal{L}_{\mathrm{KL}},
\;
\theta^{*}=\theta .
\label{eq:distillation_loss}
\end{align}

Thus, only CIVIC-specific parameters $\phi$ are optimized, while the base VLM parameters $\theta$ remain fixed.

\section{Experimental}
\label{sec:experimental}

\begin{figure*}
    \centering
    \includegraphics[width=1\linewidth]{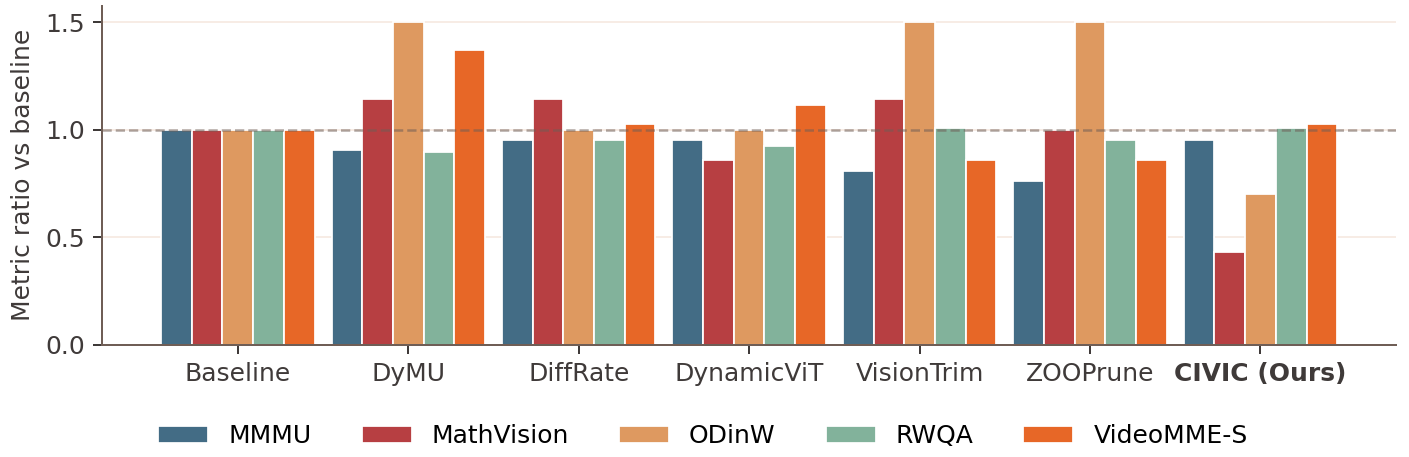}
    \caption{Relative benchmark performance of efficient VLM methods normalized by the dense baseline on MMMU, MathVision, ODinW-13, RealWorldQA (RWQA), and VideoMME short.}
    \label{fig:all_metrics}
\end{figure*}

\subsection{Experimental Setup}
\label{subsec:exp_setup}

This paper evaluates the open-source \texttt{Qwen3-VL-2B-Instruct} base model on a single NVIDIA RTX 4090 GPU to ensure reproducibility in resource-constrained environments. Multi-modal competency is evaluated across five benchmarks: reasoning (MMMU, MathVision), perception and localization (ODinW-13, RealWorldQA), and sequential context tracking (VideoMME short split). This paper compares CIVIC against five baseline token-reduction approaches: DyMU, DiffRate, DynamicViT (hard/soft variants), VisionTrim, and ZOO-Prune. Systemic efficacy is mapped across structural metrics (keep ratio, FLOPs, prefill length, peak KV cache) and physical metrics (end-to-end latency, compression overhead, throughput). To ensure timing consistency, all autoregressive decoding operations use deterministic parameters ($\tau = 0$, $\text{top-}k = 1$, $\text{top-}p = 1$) via PyTorch and Transformers under default package configurations.

\subsection{Task Accuracy Evaluation}
\label{subsec:main_accuracy}

As illustrated in Figure~\ref{fig:all_metrics}, CIVIC establishes stable performance boundaries across diverse domains when normalized against the unpatched dense baseline. Prior post-hoc pruning frameworks exhibit severe degradation and performance cliffs on reasoning-heavy tasks (MathVision) and fine-grained localization (ODinW-13) due to destructive token eviction, which irreparably fragments visual context. Conversely, CIVIC demonstrates exceptional cross-benchmark resilience, surprisingly enhancing localization accuracy. This insight reveals that learning a natively compact representation via adaptive aggregation captures global spatial relationships far better than heuristic patch dropping. It proves that preserving semantic integrity at the source eliminates the need for latency-inducing dense interface restorations during text generation.

\subsection{System Efficiency Benchmarking}
\label{subsec:practical_efficiency}

\begin{figure}[h]
    \centering
    \includegraphics[width=0.6\linewidth]{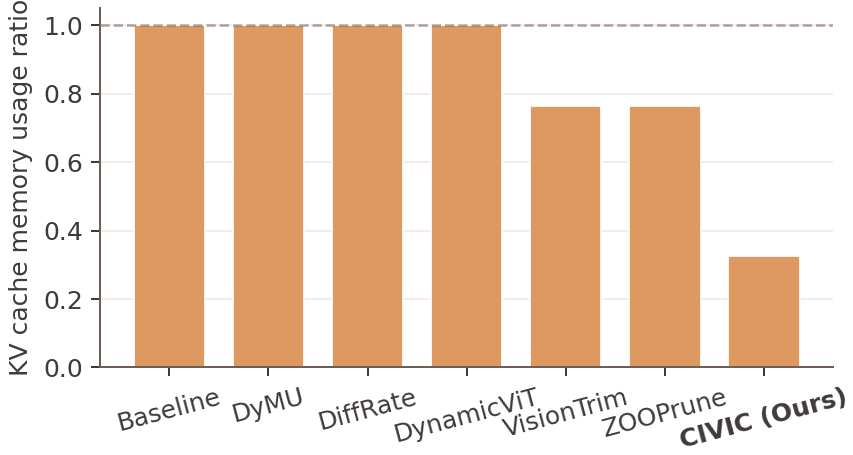}
    \caption{Relative KV-cache memory utilization across token reduction configurations.}
    \label{fig:kv_cache}
\end{figure}

Theoretical visual FLOP reductions frequently fail to capture physical deployment costs. As quantified in Table~\ref{tab:latency_breakdown}, post-hoc compression methods minimize visual tokens but leave downstream sequences uncompressed, failing to yield systemic latency reductions. CIVIC structurally optimizes the entire multimodal pathway by condensing the sequence entering the language model prefill stage. Consequently, Figure~\ref{fig:kv_cache} demonstrates that CIVIC monotonically reduces relative KV cache memory utilization to nearly one-third of the baseline allocation. Crucially, this substantial memory relief fundamentally shifts the hardware memory-compute bound, allowing for significantly larger batch sizes and extended context windows in practical deployment scenarios.

\begin{table*}[t]
\centering
\resizebox{\textwidth}{!}{
\begin{tabular}{lccccccccc}
\toprule
 \textbf{Model} & \textbf{Total Latency} $\downarrow$ & \textbf{Vision Enc.} $\downarrow$ & \textbf{Proj.} $\downarrow$ & \textbf{Prefill} $\downarrow$ & \textbf{Decode} $\downarrow$ & \textbf{LLM Total} $\downarrow$ & \textbf{Overhead} $\downarrow$ & \textbf{Prefill Tokens} $\downarrow$ & \textbf{KV Cache} $\downarrow$ \\
  & (ms) & (ms) & (ms) & (ms) & (ms) & (ms) & (ms) & (tokens) & (MB) \\
\midrule
 Baseline& 3543.0 & 47.82 & 0.39 & 35.20 & 3128.4 & 3274.3 & 0.00 & 1122.2 & 122.7 \\
 DyMU& 3688.2 & 96.74 & 0.40 & 34.04 & 3224.2 & 3258.2 & 9.28          & 1122.2 & 122.7 \\
 DiffRate& 3804.9 & 115.9 & 0.42 & 33.95 & 3311.2 & 3345.1 & 1.26          & 1122.2 & 122.7 \\
 DynamicViT (hard shrink) & 3785.3 & 52.98 & 0.39 & 34.43 & 3346.6 & 3381.1 & 3.45          & 1122.2 & 122.7 \\
 DynamicViT (soft gumbel) & 4132.6 & 186.5 & 0.39 & 34.51 & 3543.5 & 3578.0 & 2.62          & 1122.2 & 122.7 \\
 VisionTrim& 4207.5 & 47.71 & 0.38 & 26.40 & 3734.9 & 3761.3 & 3.96          & 848.5  & 92.81 \\
 ZOO-Prune& 3564.3 & 47.80 & 0.38 & 27.20 & 3141.2 & 3168.4 & 18.45         & 848.5  & 92.81 \\
 CIVIC& \textbf{2514.9} & \textbf{32.53} & \textbf{0.16} & \textbf{18.94} & \textbf{2229.4} & \textbf{2248.3} & \textbf{0.49}          & \textbf{407.9}  & \textbf{44.61} \\
\bottomrule
\end{tabular}
}
\caption{Detailed latency decomposition and sequence footprint across different multi-modal inference stages, where bold values indicate the best performance in each metric. For DynamicViT, \textit{hard shrink} denotes a deterministic token dropping strategy based on score ranking, while \textit{soft gumbel} incorporates a stochastic Gumbel-Softmax sampling layer to learn token routing probabilities.}
\label{tab:latency_breakdown}
\end{table*}

\begin{figure}[h]
    \centering
    \includegraphics[width=0.6\linewidth]{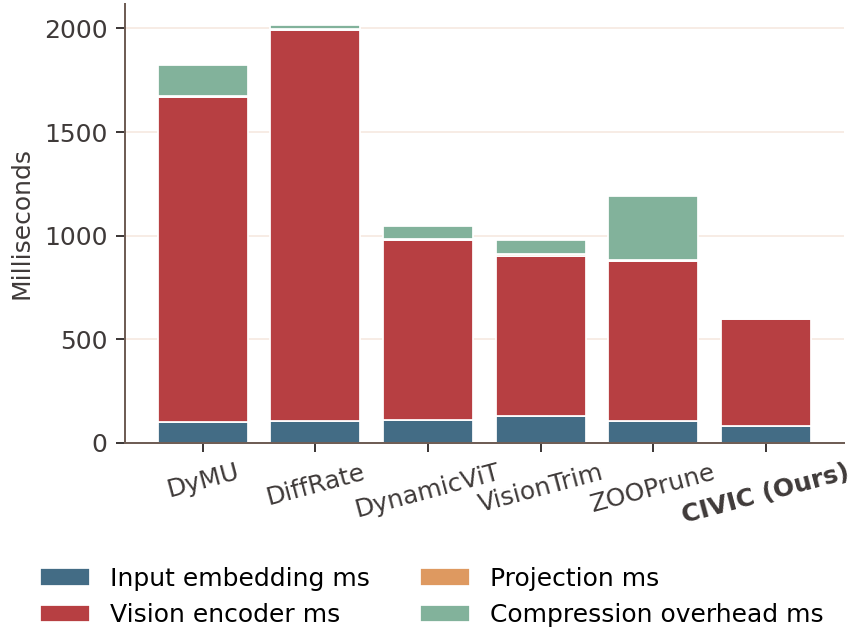}
    \caption{Granular wall-clock execution latency (ms) decomposed across pipeline stages. The path-consistent framework of \texttt{CIVIC} reduces attention and decoding durations without introducing significant compression logic bottlenecks.}
    \label{fig:timing_decomposition}
\end{figure}

\subsection{The Theory-Practice Efficiency Gap}
\label{subsec:gap_analysis}

Figure~\ref{fig:gap_analysis_summary} exposes a pronounced discrepancy between idealized token reduction and practical runtime. Baseline frameworks report high theoretical pruning ratios but suffer severe latency regressions. This paradox highlights the illusion of FLOP counting: operations like dynamic feature scoring, localized routing, and non-contiguous gather/scatter memory access fundamentally break GPU hardware efficiency. As quantified in Table~\ref{tab:latency_breakdown}, ZOOPrune and DyMU incur heavy computational bookkeeping overheads of 18.45 ms and 9.28 ms, respectively, stalling execution and neutralizing theoretical savings. By preserving a static, compact representation natively within the primary inference graph, CIVIC maintains memory contiguity. It eliminates runtime token eviction, yielding a negligible compression overhead of 0.49 ms and ensuring theoretical reductions directly drive physical wall-clock acceleration.

\begin{table*}[t]
\centering
\resizebox{\textwidth}{!}{
\begin{tabular}{llcccccc}
\toprule
\textbf{Ablation Group} & \textbf{Variant} & \textbf{Total (ms)} $\downarrow$ & \textbf{Vision Enc. (ms)} $\downarrow$ & \textbf{Proj. (ms)} $\downarrow$ & \textbf{Prefill (ms)} $\downarrow$ & \textbf{Decode (ms)} $\downarrow$ & \textbf{Overhead (ms)} $\downarrow$ \\
\midrule
Compact Tokens & C=64 & 2823.7 & 30.2 & \textbf{0.12} & \textbf{15.6} & 2514.4 & 4.92 \\
 & C=256 & 2440.8 & 40.2 & 0.21 & 22.2 & 2151.3 & 5.13 \\
 & C=512 & \textbf{2327.6} & 45.9 & 0.27 & 26.0 & \textbf{2038.0} & 5.43 \\
\midrule
Min Keep Ratio & Min=0.0 & 2591.8 & 33.9 & 0.16 & 18.3 & 2293.2 & 4.99 \\
 & Min=0.2 & 2635.8 & 34.0 & 0.16 & 18.3 & 2334.0 & 4.97 \\
 & Min=0.5 & 2616.6 & 34.1 & 0.16 & 18.4 & 2321.1 & 4.97 \\
\midrule
KV Compression & No KV & 2598.9 & \textbf{23.8} & 0.16 & 18.4 & 2314.4 & \textbf{4.90} \\
 & Anchors=128 & 2543.4 & 33.1 & 0.16 & 18.3 & 2252.8 & 5.02 \\
 & Anchors=512 & 2660.3 & 36.9 & 0.16 & 18.4 & 2356.4 & 4.95 \\
\bottomrule
\end{tabular}
}
\caption{Granular stage-wise execution latency across architectural ablation variants (best values in bold). Modifying structural constraints like token budget (C) or attention mechanisms systematically redistributes computational load across the sequential stages of the inference pipeline.}
\label{tab:ablation_latency}
\end{table*}

\begin{figure*}[h]
    \centering
    \includegraphics[width=1\linewidth]{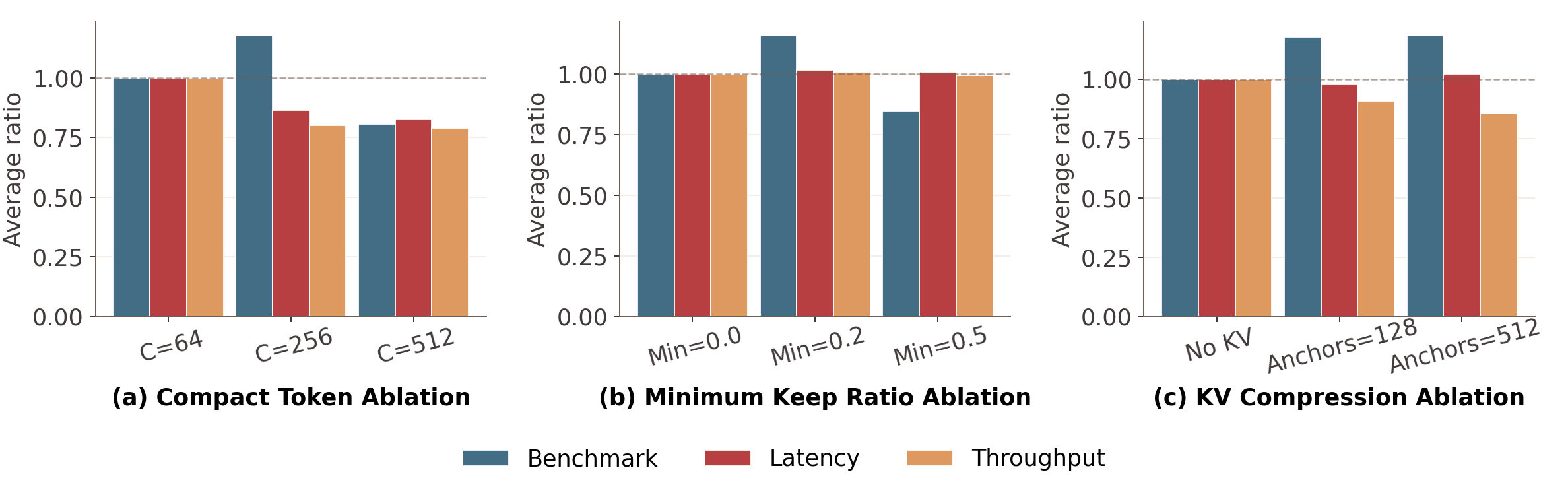}
    \caption{Ablation analysis of CIVIC's core architectural axes. \textbf{(a) Compact Token Budget:} Illustrates the Pareto frontier between visual information density and downstream sequence prefill workloads. \textbf{(b) Adaptive Minimum Keep Ratio:} Demonstrates the retention floor's role in safeguarding fine-grained visual features against aggressive truncation. \textbf{(c) KV-Compressed Vision Attention:} Benchmarks physical execution speed against memory allocation overhead across different key-value anchor configurations.}
    \label{fig:ablation_metrics}
\end{figure*}

\subsection{Granular Latency Decomposition}
\label{subsec:timing_breakdown}

Shown in Figure~\ref{fig:timing_decomposition}, Table~\ref{tab:latency_breakdown}, disaggregating execution latency across pipeline stages reveals critical insights into where computational bottlenecks shift. Conventional methods confine sequence reductions strictly to the visual encoder, forcing the downstream language model to process uncompressed representations (1122.2 prefill tokens). Because every generated text token must attend to the entire visual prompt, this uncompressed context structurally bottlenecks the autoregressive decoding phase, which dominates over 90\% of total execution time. 

In contrast, CIVIC propagates a structurally shortened stream directly through the projection layer (0.16 ms) and into the language model, cutting the prefill token count to 407.9. This uninterrupted, end-to-end path consistency triggers cascading latency relief. By drastically reducing the cross-attention burden for every subsequent generated token, CIVIC shortens prefill execution to 18.94 ms and slashes autoregressive decoding to 2229.4 ms. Ultimately, CIVIC minimizes cumulative execution to 2248.3 ms, achieving the lowest total pipeline latency of 2514.9 ms and proving that holistic sequence compression is the definitive driver of tangible VLM acceleration.

\subsection{Ablation Study}
\label{subsec:ablation_study}

Ablation experiments on the Qwen3-VL base model strictly isolate architectural variables using identical distillation and hardware profiling. This study systematically evaluates the \textit{token budget}, the maximum visual tokens propagated downstream, the \textit{keep ratio}, the minimum proportion of spatially relevant tokens preserved, and \textit{KV-compressed vision attention}, the structural reduction of key-value caches to minimize memory allocation. Table~\ref{tab:ablation_latency} and Figure~\ref{fig:ablation_metrics} validate these components:

\textbf{Compact Token Budget:} Testing capacities reveal that extreme compression (C=64) minimizes prefill latency but starves context, paradoxically inflating autoregressive decoding time as the model struggles with fragmented information. Conversely, expanded budgets (C=512) bottleneck early vision stages. Intermediate budgets optimally balance visual density with downstream generative workloads.

\textbf{Adaptive Minimum Keep Ratio:} Unbounded eviction (Min=0.0) yields negligible speedups while destroying high-detail visual accuracy. A conservative retention floor is thus critical for safeguarding spatial integrity against overly aggressive pruning heuristics.

\textbf{KV-Compressed Vision Attention:} Native attention compression achieves true end-to-end acceleration by preventing sequential memory fragmentation and minimizing computational bookkeeping. This proves that physical execution efficiency relies on contiguous memory patterns rather than theoretical FLOP reductions alone.

\textbf{Compact Multimodal Prefill:} Artificially restoring dense placeholders prior to the language model instantly nullifies all speedups. This exposes a core fallacy of baseline methods: intermediate visual compactness is functionally futile unless it persists seamlessly into the LLM.

\textbf{Text-Aligned KL Distillation:} Replacing standard distillation with a text-aligned KL objective prevents severe generative degradation. This ensures the compressed visual manifold maps strictly to expected target distributions, successfully bridging the semantic gap between condensed vision features and the frozen language space.

\section{Conclusion}
This paper addresses systemic inefficiencies where theoretical token reductions fail to yield physical VLM acceleration. Because conventional post-hoc pruning bottlenecks inference via dense restorations, CIVIC enforces path consistency, maintaining compact sequences from visual ingestion through autoregressive decoding. Empirical results demonstrate that CIVIC reduces the prefill sequence length from 1122.2 to 407.9 tokens, scales down KV-cache memory utilization from 122.7 MB to 44.61 MB, and decreases total inference latency from 3543.0 ms to 2514.9 ms while preserving performance across benchmarks. These metrics validate that holistic sequence compactness successfully translates theoretical compression into physical hardware efficiency. Future research will explore extending this path-consistent design to dynamic, instance-adaptive token budgets and validating its scalability across multi-image contexts and long-form video streams.

\section{Limitations}
Current evaluations rely on static token budgets and are restricted to single images using the Qwen3-VL-2B architecture. Future research must validate scalability across larger models, dynamic instance-adaptive capacities, and long-form video or multi-image contexts. 

\section{Ethical Considerations}
While deploying efficient VLMs on edge devices provides real-time benefits, offline inference bypasses traditional server-side guardrails. This amplifies dual-use risks necessitating the concurrent development of robust on-device safety mechanisms and responsible deployment guidelines.

\section{Acknowledgments}
We acknowledge the use of large language models strictly as writing assistants for grammatical polishing and structural refinement during the preparation of this manuscript

\bibliographystyle{unsrtnat}
\bibliography{1-references}  






\end{document}